\documentclass{article}

\PassOptionsToPackage{numbers, compress}{natbib}

\usepackage[final]{neurips_2023}

\usepackage[utf8]{inputenc} 
\usepackage[T1]{fontenc}    
\usepackage{hyperref}       
\usepackage{url}            
\usepackage{booktabs}       
\usepackage{amsfonts}       
\usepackage{nicefrac}       
\usepackage{microtype}      
\usepackage{xcolor}         

\usepackage{subcaption}
\usepackage{graphicx}
\usepackage{subcaption}
\usepackage{amsmath}
\title{Diffusing More Objects for Semi-Supervised\\Domain Adaptation with Less Labeling}

\author{%
  Leander van den Heuvel \\
  TNO Netherlands\\
  \And
  Gertjan Burghouts \\
  TNO Netherlands\\
  \AND
  David W. Zhang \\
  University of Amsterdam \\
  \And
  Gwenn Englebienne \\
  University of Twente \\
  \And
  Sabina B. van Rooij \\
  TNO Netherlands \\
}
\DeclareMathOperator{\nms}{nms}
\DeclareMathOperator{\concat}{concat}

\begin{document}

\maketitle

\begin{abstract}
  For object detection, it is possible to view the prediction of bounding boxes as a reverse diffusion process. Using a diffusion model, the random bounding boxes are iteratively refined in a denoising step, conditioned on the image. We propose a stochastic accumulator function that starts each run with random bounding boxes and combines the slightly different predictions. We empirically verify that this improves detection performance. The improved detections are leveraged on unlabelled images as weighted pseudo-labels for semi-supervised learning. We evaluate the method on a challenging out-of-domain test set. Our method brings significant improvements and is on par with human-selected pseudo-labels, while not requiring any human involvement.
\end{abstract}

\section{Introduction}

Diffusion models achieved success on a wide range of generative tasks, including synthetic images, see e.g. recent surveys \citep{cao2022survey, yang2022diffusion}. Recently, they have been explored for other tasks, such as image segmentation \citep{baranchuk2021label} and object detection \citep{Chen2022}. DiffusionDet \citep{Chen2022} formulates object detection as a box denoising process conditioned on the image input, starting with a set of random bounding boxes that is iteratively improved (denoised). This is advantageous because it avoids the necessity of region proposals \citep{he2017mask}, anchors \citep{lin2017focal}, or learnable queries \citep{zhu2020deformable}. The process of denoising object boxes is stochastic since it depends on the random initialization of the boxes. This brings the advantage that each run leads to slightly different outputs. Running the model multiple times may increase the probability of detecting objects that are difficult to detect. We hypothesize that this is especially useful when there is a significant domain gap between the training distribution and test distribution. In this work we aim to solve the domain gap caused by different viewpoints: from frontal view (source domain) to aerial view (target domain). To that end, we propose a stochastic accumulator of the diffusion model's outputs to acquire improved object detections in the target domain. We anticipate that the improved detections are more useful as pseudo-labels for semi-supervised learning. For that purpose, we reformulate the diffusion model's loss function to include unlabeled images and their pseudo-labels. Since the pseudo-labels are imperfect, we propose a weighted loss that takes their confidence into account. Our contribution is two-fold: we propose a stochastic accumulator and a semi-supervised weighted pseudo-label loss. We show their efficacy on a challenging domain gap from frontal everyday images (MS-COCO \citep{Lin2014MicrosoftContext}) to aerial images that cover day and night and were taken from much larger distances (VisDrone \citep{Zhu2020DetectionChallenge}). We demonstrate that our method is on par with human-selected pseudo-labels.
\section{Related Work}
Aiming at label-efficient domain adaptation for object detection, we consider related work in domain adaptation and semi-supervised object detection. For domain adaptation, there are unsupervised e.g. \cite{Liu2022DeepPerspectives,Oza} and supervised training procedures e.g. \cite{Motiian,Wang2018}. With supervision in the target domain, adaptation methods can improve greatly \cite{Oza}, however, supervised training requires labeled data. Semi-supervised training uses both labeled and unlabeled images \cite{Yang2022ALearning}. One approach is the use of pseudo-labels on the unlabeled images. In \citet{Sohn2020ADetection} a self-training and the augmentation driven consistency regularization (STAC) for object detection is proposed. This work introduces a trained teacher model that generates pseudo-labels, which are used to finetune a student model. The authors note that class imbalance of the labeled images could lead to reduced pseudo-label quality. To address this issue, Unbiased Teacher is proposed by \citet{Liu2021}, that uses the Focal loss \cite{Lin2017} to down-weigh overly confident pseudo-labels. STAC and Unbiased Teacher show impressive results for training with fewer labels. However, both papers focus on in-domain performance improvements using pseudo-labels. We explore semi-supervised training with pseudo-labels for domain adaptation.
\section{Method}

We extend DiffusionDet \citep{Chen2022} (Section \ref{diffdet}) in two ways. First, we improve its predictions for a large domain gap, using a stochastic accumulator (Section \ref{stoch}). Secondly, we use the improved predictions as weighted pseudo-labels for semi-supervised learning (Section \ref{semisup}).

\begin{figure}[!h]
    \centering
    \includegraphics[width=0.99\textwidth]{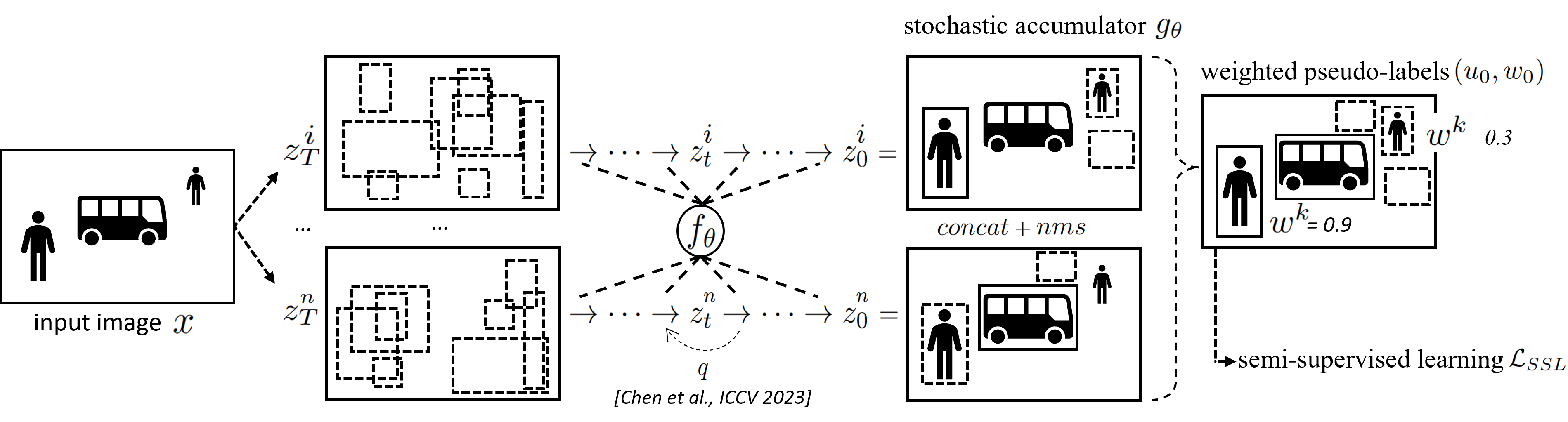}
    \caption{Our method improves object detection by exploiting the stochastic nature of the diffusion-based model, thereby accumulating diverse predictions, and leveraging them as weighted pseudo-labels in semi-supervised learning.}\label{fig:method}
\end{figure}

\subsection{Preliminaries}\label{diffdet}

DiffusionDet \citep{Chen2022} trains a neural network $f_\theta(z_t, t)$ to predict a set of boxes $z_0$ from noisy boxes $z_t$ by minimizing a combination of losses including the $\ell_2$ loss: $\mathcal{L} = \frac{1}{2}||f_\theta(z_t, t) - z_0||^2$. The boxes are a set of coordinates in the image, $z \in \mathbf{R}^{t \times 4}$, with for each box a category label. The noisy boxes $z_t$ are taken from a forward noise process that gradually transforms the boxes $z_0$ to noisy versions $z_t$: $q(z_t|z_0) = \mathcal{N}((z_t|\sqrt{\overline{\alpha_t}})z_0, (1 - \overline{\alpha_t}) \textit{\textbf{I}})$, where $\overline{\alpha_t}$ defines the noise schedule. At inference, $z_0$ is predicted from noise $z_T$ by $f_\theta$ iteratively, $z_T \rightarrow \dots \rightarrow z_t \rightarrow \dots \rightarrow z_0$. The network $f_\theta(z_t, t, x)$ is trained to predict $z_0$ from noisy boxes $z_t$ for an image $x$. For more details, we refer to \cite{Chen2022}.

\subsection{Stochastic accumulator}\label{stoch}

The noise $z_T$ is a random variable. Hence, we can apply $f_\theta$ for an image $x$ with a different draw $i$ of noisy bounding boxes, $z_T^i$, leading to a prediction $z_0^i = f_\theta(z_T^i, t, x)$. We can do this multiple times, each time leading to a (slightly) different prediction, $\{z_0^i\}_{i \in 1 .. n}$. The rationale is that the object detection can be improved by accumulating the model's outputs over these multiple runs. Our accumulator function is: 

\begin{equation}\label{eq:accumulator}
    g_\theta(x) \,\, = \,\, \nms(\concat(\{f_\theta(z_T^i, t, x)\}_{i \in 1 .. n}))
\end{equation}\\
where $\concat(\cdot)$ is the operator that concatenates the outputs of $f_\theta$ in a row-wise manner, and $\nms(\cdot)$ is the non-maximum-suppression operator. The resulting prediction $(u_0, w_0) = g_\theta(x)$ is a set of boxes $u$ with respective confidences $w$. This procedure is illustrated in Figure \ref{fig:method}.

In case of large domain gaps, where objects from the train and test set differ in e.g. viewpoint, size, or background, some objects in the test set may be difficult to detect. With $g_\theta(x)$, we leverage the different, stochastic outputs of the model between runs, thereby increasing the probability of finding challenging objects in the test set. 


\subsection{Weighted pseudo-labels for semi-supervised learning}\label{semisup}

The rationale is to use the enhanced object predictions $(u_0, w_0) = g_\theta(x)$ as pseudo-labels in a semi-supervised setup. This is done to further improve the overall model $g_\theta$ on the target domain, by finetuning its predictor function $f_\theta$. We assume that for some images the objects are labeled, $z_0$, while for others no labels are available. For the unlabeled images, we have the predicted boxes and their confidence, i.e., $(u_0, w_0)$. We reformulate the original loss of DiffusionDet, which was fully supervised (see Section \ref{diffdet}), as a loss that enables semi-supervised learning (SSL) over $m$ labeled images and $l$ unlabelled images:

\begin{equation}
    \mathcal{L}_{SSL} = \frac{\frac{1}{2} \sum_{k=1}^{m} ||f_\theta(z_t^k, t) - z_0^k||^2}{m} + \frac{\frac{1}{2} \sum_{k=1}^{l} ||f_\theta(z_t, t) - u_0^k||^2 \cdot w^k}{\sum_{k=1}^{l} w^k}
\end{equation}

Our semi-supervised approach to reduce the domain gap is illustrated in the right part of Figure \ref{fig:method}.
\section{Experiments}

\subsection{Setup and domain gap}

\begin{figure}[!h]
     \centering
     \begin{subfigure}[b]{\textwidth}
         \centering
         \includegraphics[width=0.49\textwidth]{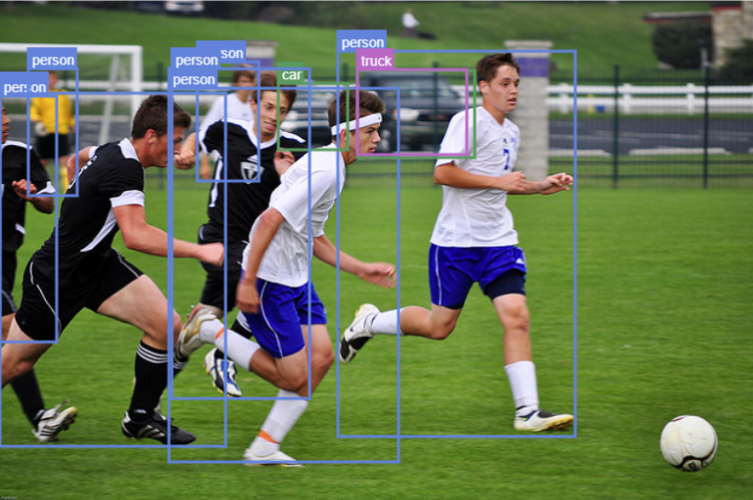}
         \includegraphics[width=0.49\textwidth]{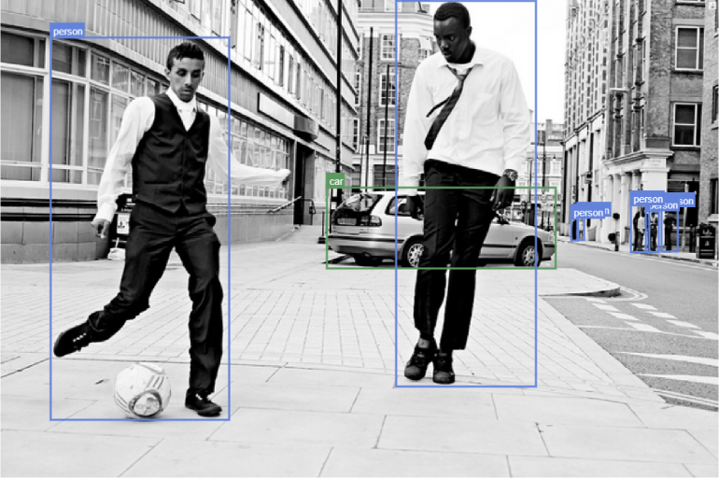}
         \caption{Source domain: MS-COCO \cite{Lin2014MicrosoftContext}}
     \end{subfigure}\\
     \centering
     \begin{subfigure}[b]{\textwidth}
         \centering
         \includegraphics[width=0.49\textwidth]{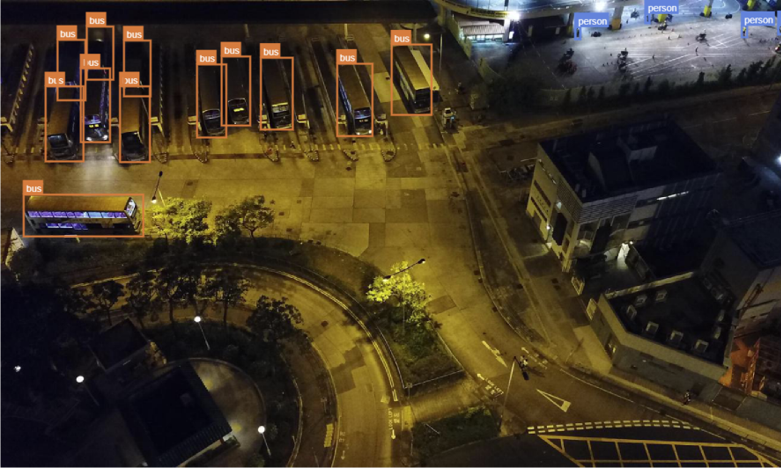}
         \includegraphics[width=0.49\textwidth]{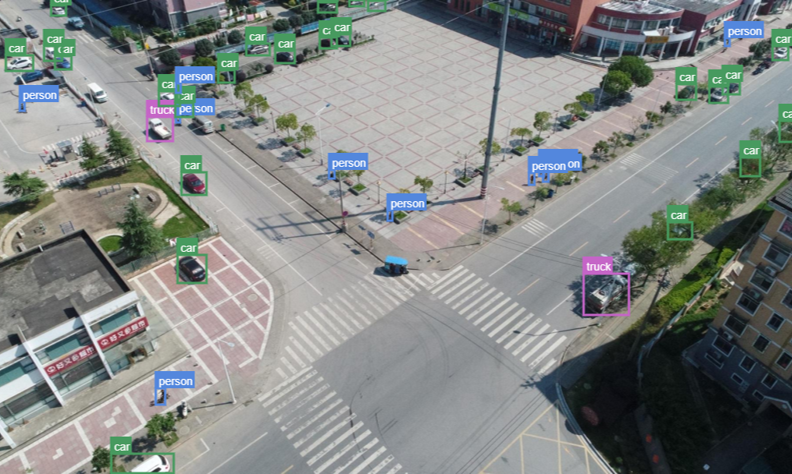}
         \caption{Target domain: VisDrone \cite{Zhu2020DetectionChallenge}}
     \end{subfigure}
     \caption{Examples from the source (a) and target (b) domains. The target domain is very different, with changes in viewpoint, day and night, and distance to the objects.}
     \label{fig:datasets}
\end{figure}

The source domain is MS-COCO \citep{Lin2014MicrosoftContext} with everyday frontal images. The target domain is VisDrone \citep{Zhu2020DetectionChallenge} with aerial images taken by drones in urban areas. Images in VisDrone are captured from a greater distance using a drone, leading to smaller objects, and, in some cases, motion blur. We are interested in a domain adaptation problem where the classes between the train and test set overlap. For evaluation, we select overlapping classes between MS-COCO and VisDrone: car, person (including pedestrian from VisDrone), bus, and truck. A fragment of an image from VisDrone is shown in Figure \ref{fig:example_data_samples} where the objects have an aerial view, are relatively small compared to MS-COCO and contain motion blur. More examples are shown in Figure \ref{fig:datasets} (Appendix). The domain gap is severe, not only in appearance, but also the distribu tion of object sizes is very different, see Figure \ref{fig:gap} (Appendix). Our baselines are YOLOv5 \citep{Jocher2022Ultralytics/yolov5:v7.0} and DiffusionDet \citep{Chen2022} with the standard settings (300 random proposal boxes and 10 reverse diffusion steps). We evaluate the merit of our extensions to DiffusionDet. 

\subsection{Object detection can be improved by stochasticity}\label{sec:exp1}

\begin{figure}[!h]
     \centering
     \begin{subfigure}[b]{0.48\textwidth}
         \centering
         \includegraphics[width=\textwidth]{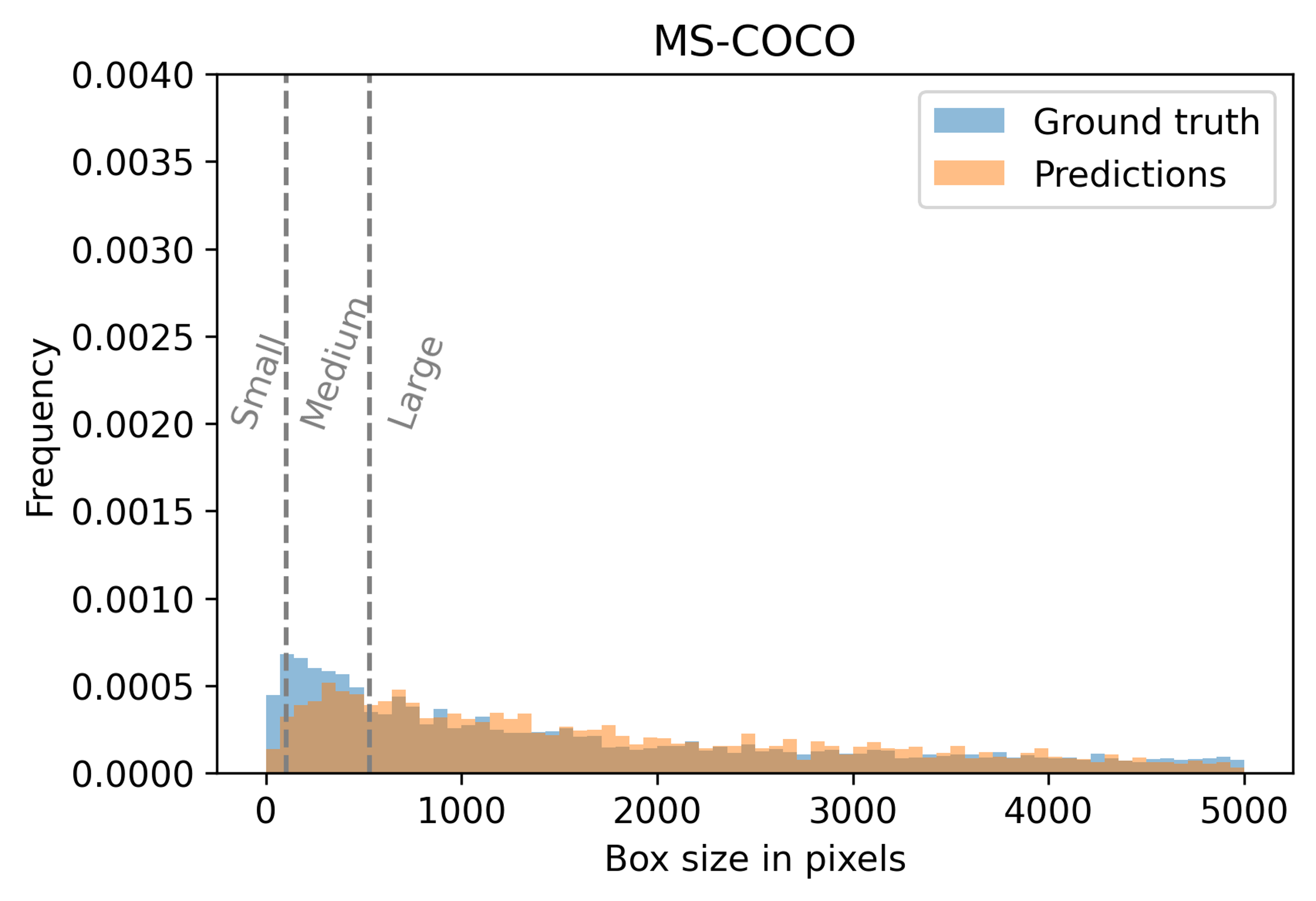}
         \caption{Source domain: MS-COCO \cite{Lin2014MicrosoftContext}}
     \end{subfigure}
     \begin{subfigure}[b]{0.48\textwidth}
         \centering
         \includegraphics[width=\textwidth]{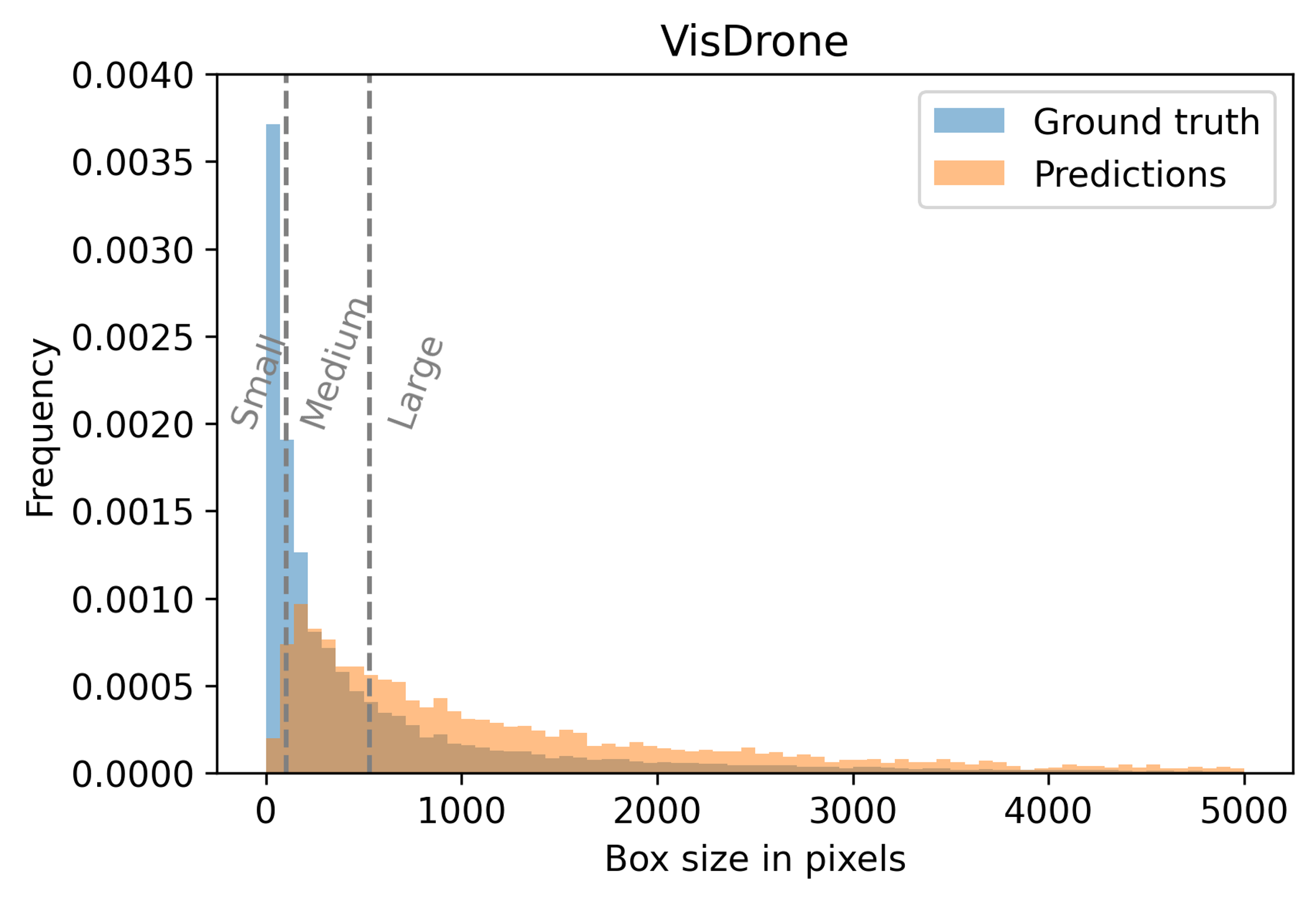}
         \caption{Target domain: VisDrone \cite{Zhu2020DetectionChallenge}}
     \end{subfigure}
     \caption{The groundtruth object sizes are distributed very differently for the source (a) and target (b) domains. The distribution of predictions by the DiffusionDet baseline do not align well with the groundtruth.}
     \label{fig:gap}
\end{figure}

Figure \ref{fig:gap} (Appendix) visualizes that the DiffusionDet baseline has problems with small objects. Small objects are highly prevalent in the target domain, but they are less frequently predicted by the baseline. This is also concluded from the performance metrics (mAP) in Table \ref{tab:stochastic}, where we evaluate the performance of various models on different object sizes in the target domain. The small objects are very difficult to detect: YOLOv5 and DiffusionDet respectively have a random or almost random performance. The best improvement over the baseline is achieved by running our accumulator for $n$ = 18 runs (Equation \ref{eq:accumulator}). It increases the mAP from 0.02 to 0.09 for small objects, and from 0.18 to 0.38 for large objects. 

\begin{figure}[!h]
     \centering
     \begin{subfigure}[b]{0.49\textwidth}
         \includegraphics[width=\textwidth]{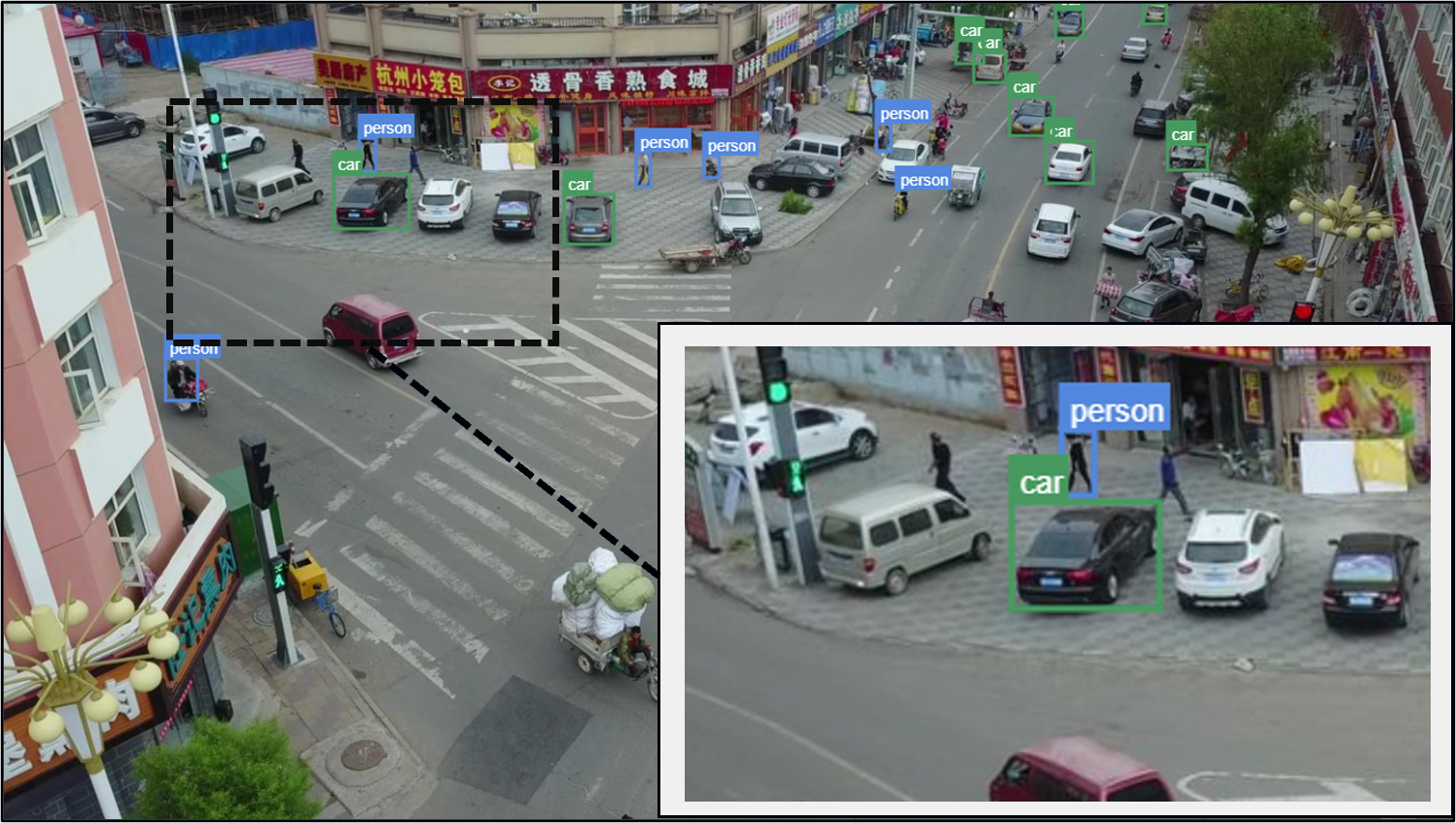}
         \caption{DiffusionDet}
     \end{subfigure}
     \begin{subfigure}[b]{0.49\textwidth}
         \includegraphics[width=\textwidth]{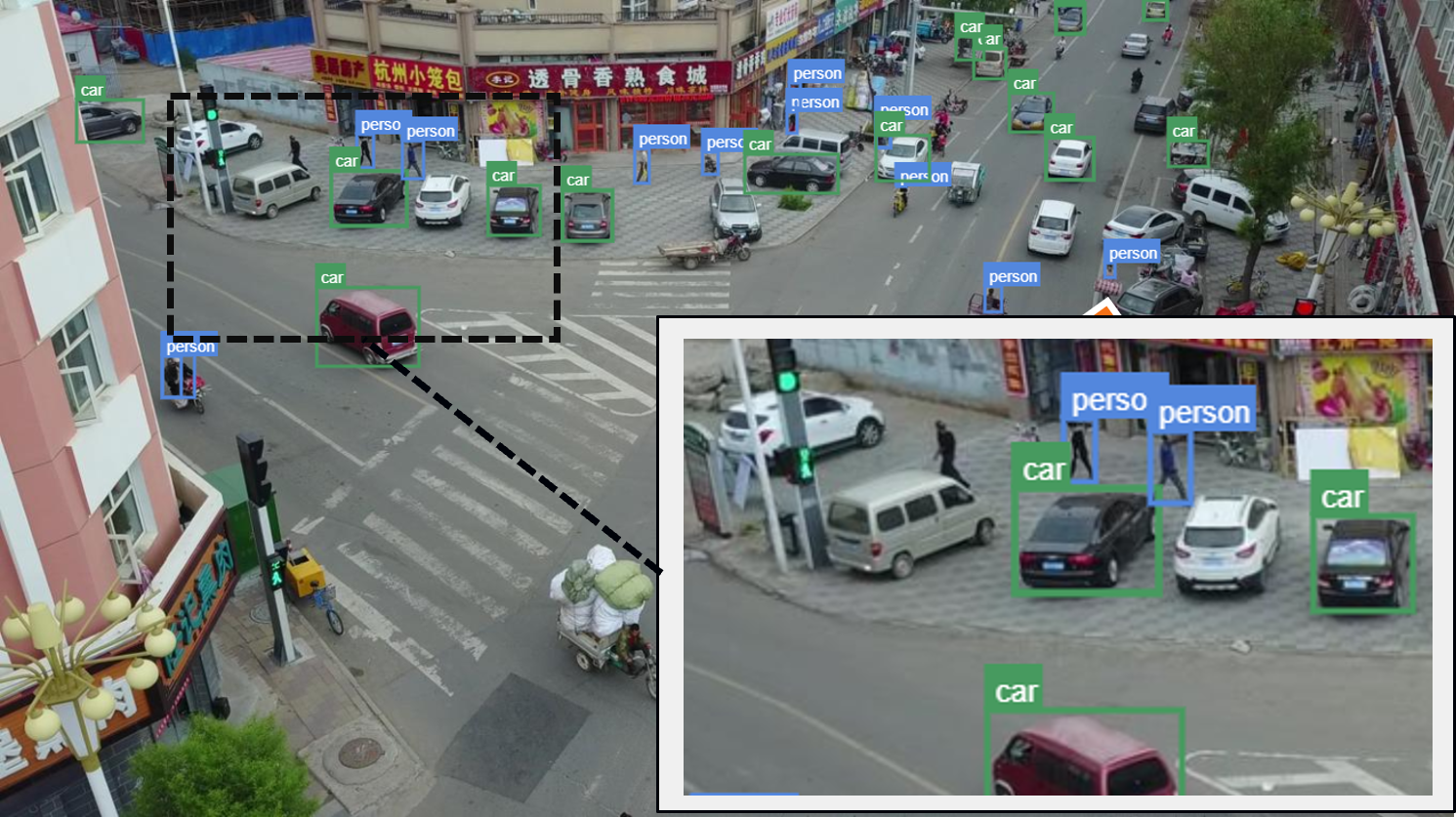}
         \caption{Stochastic accumulator, $n$ = 2}
     \end{subfigure}\\
     \centering
     \begin{subfigure}[b]{0.49\textwidth}
         \includegraphics[width=\textwidth]{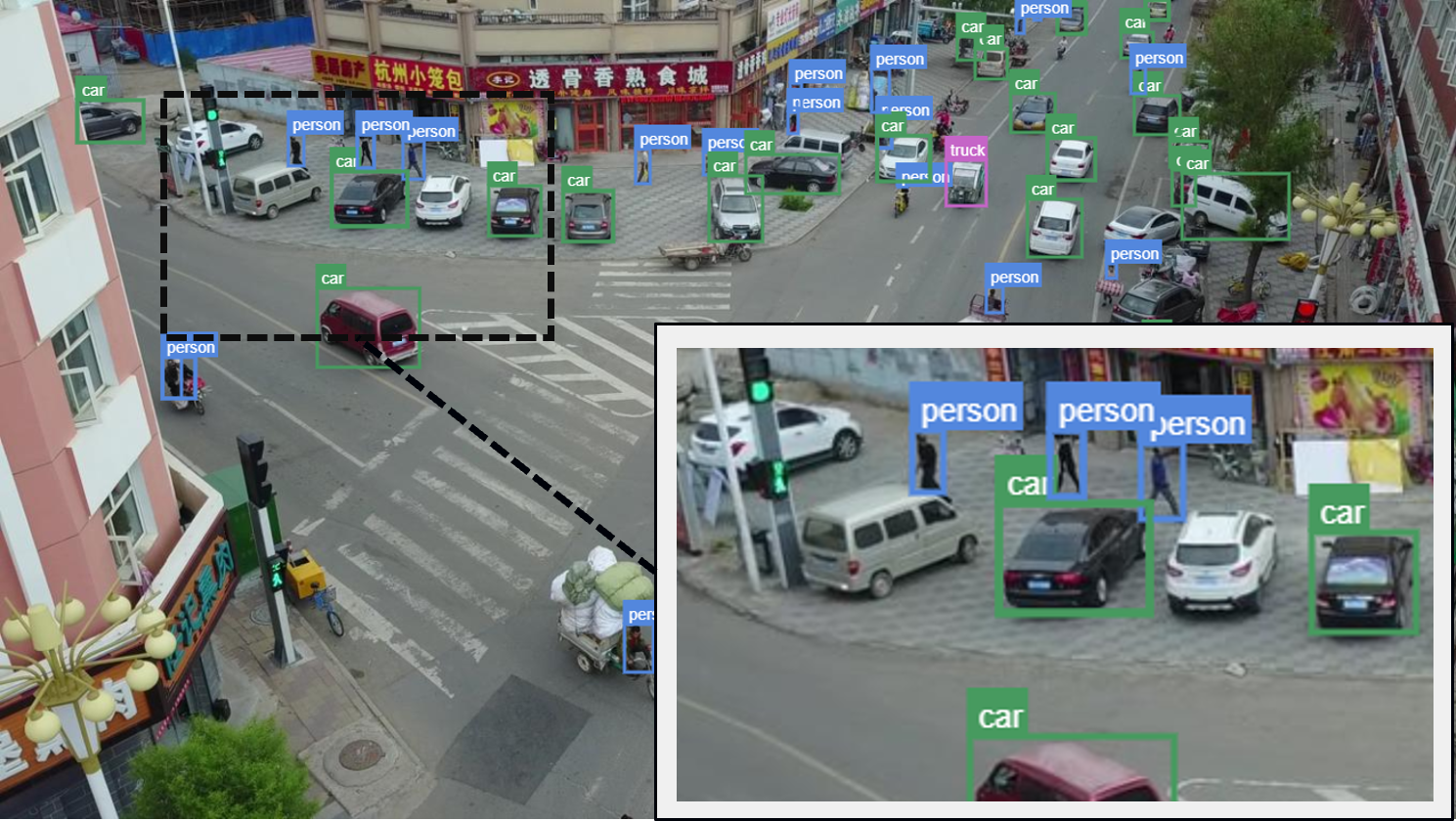}
         \caption{Stochastic accumulator, $n$ = 3}
     \end{subfigure}
     \begin{subfigure}[b]{0.49\textwidth}
         \includegraphics[width=\textwidth]{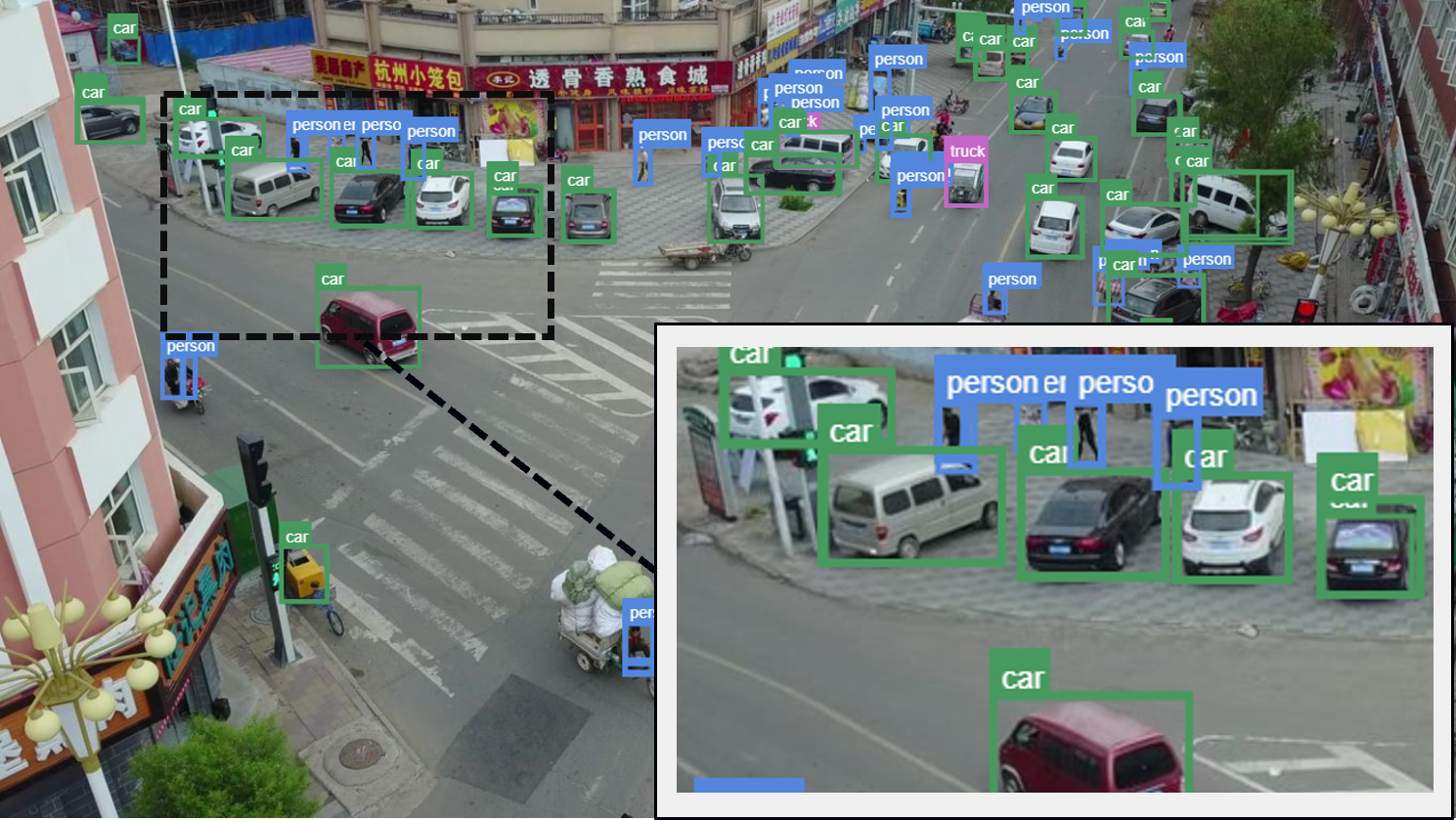}
         \caption{Stochastic accumulator, $n$ = 18}
     \end{subfigure}
     \caption{Improving DiffusionDet with our stochastic accumulator, with various runs $n$ (Equation \ref{eq:accumulator}).}
     \label{fig:stoch_impr}
\end{figure}

Additionally, we experimented with two other adjustments to the stochastic input of DiffusionDet: increasing the number of input boxes and reducing the input box sizes to align with the smaller objects in the target domain. These adjustments are less effective; reducing the target boxes to half of the size ($f_{box\_size}$ = 0.5) degrades the results; more boxes (5400) are effective but not as much as our extension. Here, 5400 boxes correspond to 18 runs with the 300 default boxes of DiffusionDet. Our stochastic accumulator is effective in improving performance under the tested large domain gap. Illustrations of improvements over the baseline are shown for various $n$ in Figure \ref{fig:stoch_impr} (Appendix). We show that, in line with our hypothesis, the diffusion model's stochasticity can be exploited to detect more objects.

\begin{table}[!h]
\centering
\resizebox{\linewidth}{!}{%
\tiny
\begin{tabular}{l ccc ccc ccc c}
\toprule
object size & \multicolumn{3}{c}{accumulator runs ($n$) (proposed)} & \multicolumn{3}{c}{random proposal boxes} & \multicolumn{3}{c}{\text{$f_{box\_size}$}} & YOLOv5 \\ \cmidrule(lr){2-4} \cmidrule(lr){5-7} \cmidrule(lr){8-10} 
 & 1 & 9 & 18 & 300 & 2700 & 5400 & 1 & 0.75 & 0.5 &  \\ \hline
small  & 0.02 & 0.08 & \textbf{0.09} & 0.02 & 0.02 & 0.03 & 0.02 & 0.00 & 0.01 & 0.00 \\
medium & 0.06 & 0.17 & \textbf{0.18} & 0.09 & 0.13 & 0.17 & 0.09 & 0.04 & 0.05 & 0.08 \\
large  & 0.18 & 0.35 & \textbf{0.38} & 0.18 & 0.34 & 0.37 & 0.18 & 0.13 & 0.15 & 0.29 \\
\bottomrule
\end{tabular}
}
\vspace{0.1cm}
\caption{VisDrone object detection performance (mAP). Our accumulator (left) performs best, with more runs it is more effective, exploiting the stochasticity of the underlying diffusion model.
}
\label{tab:stochastic}
\end{table}

\subsection{Pseudo-labels: weighted loss vs. human-selected}

We use the improved predictions as pseudo-labels in semi-supervised learning, in combination with labels from a few (10 or 50) images. Pseudo-labels may be noisy, for which we validate the merit of our weighted loss (see Section \ref{semisup}). To provide a strong baseline, we compare against human-verified pseudo-labels generated from DiffusionDet. The human annotator has selected regions in the unlabeled images that contain good pseudo-labels, meaning that the boxes are approximately correct and the class label is correct. We validate how our weighted usage of pseudo-labels, requiring no human involvement, compares to the human-verified pseudo-labels. For completeness, we also compare against the groundtruth labels within the human-verified image regions, to compare the pseudo-labels with actual labels. 

\begin{figure}[!h]
    \begin{minipage}[b]{.4\textwidth}
    \centering
    \includegraphics[width=0.99\linewidth]{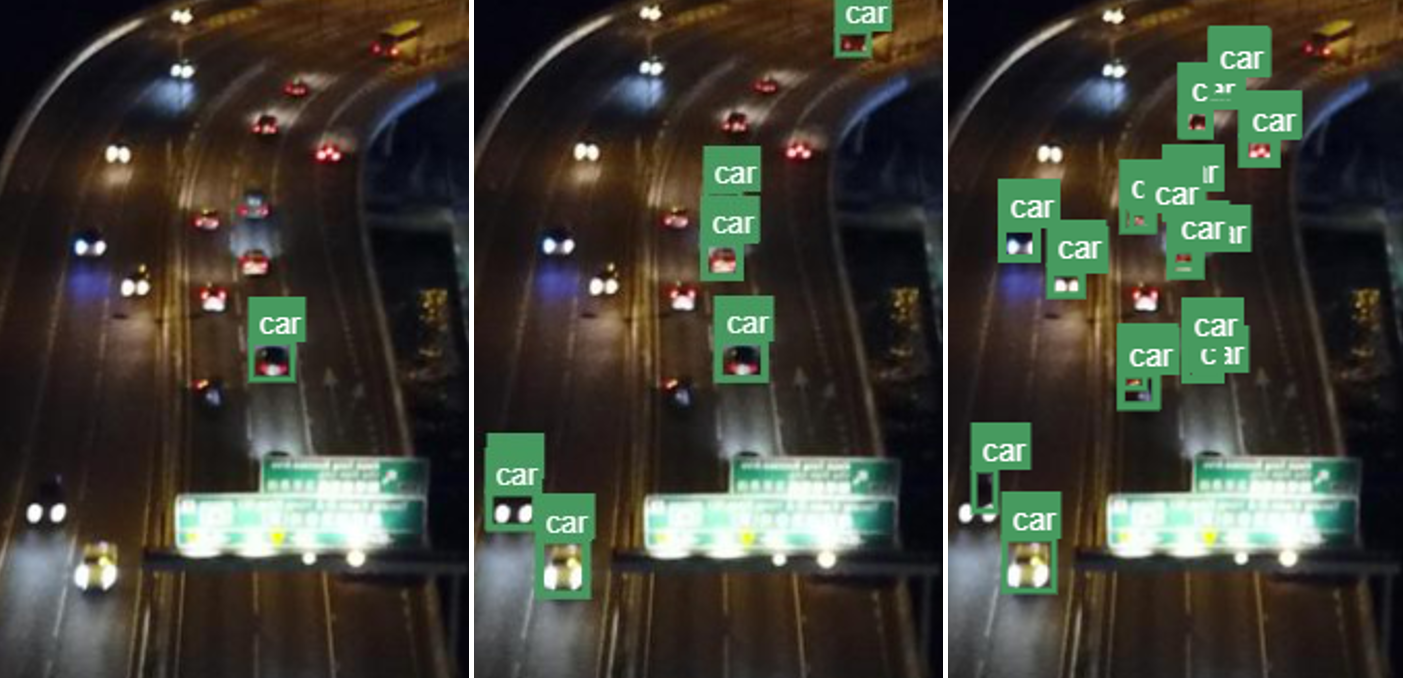}
    \captionof{figure}{
    Detections in VisDrone using DiffusionDet shown from left to right: standard model (300 boxes and 1 run), finetuned with 50 GT samples and 18 runs (baseline) , baseline with unverified weighted pseudo-labels and 18 runs}
     \label{fig:example_data_samples}
    
    \end{minipage}
    \hfill
    \begin{minipage}[b]{.58\textwidth}
    \centering
    \includegraphics[width=0.99\linewidth]{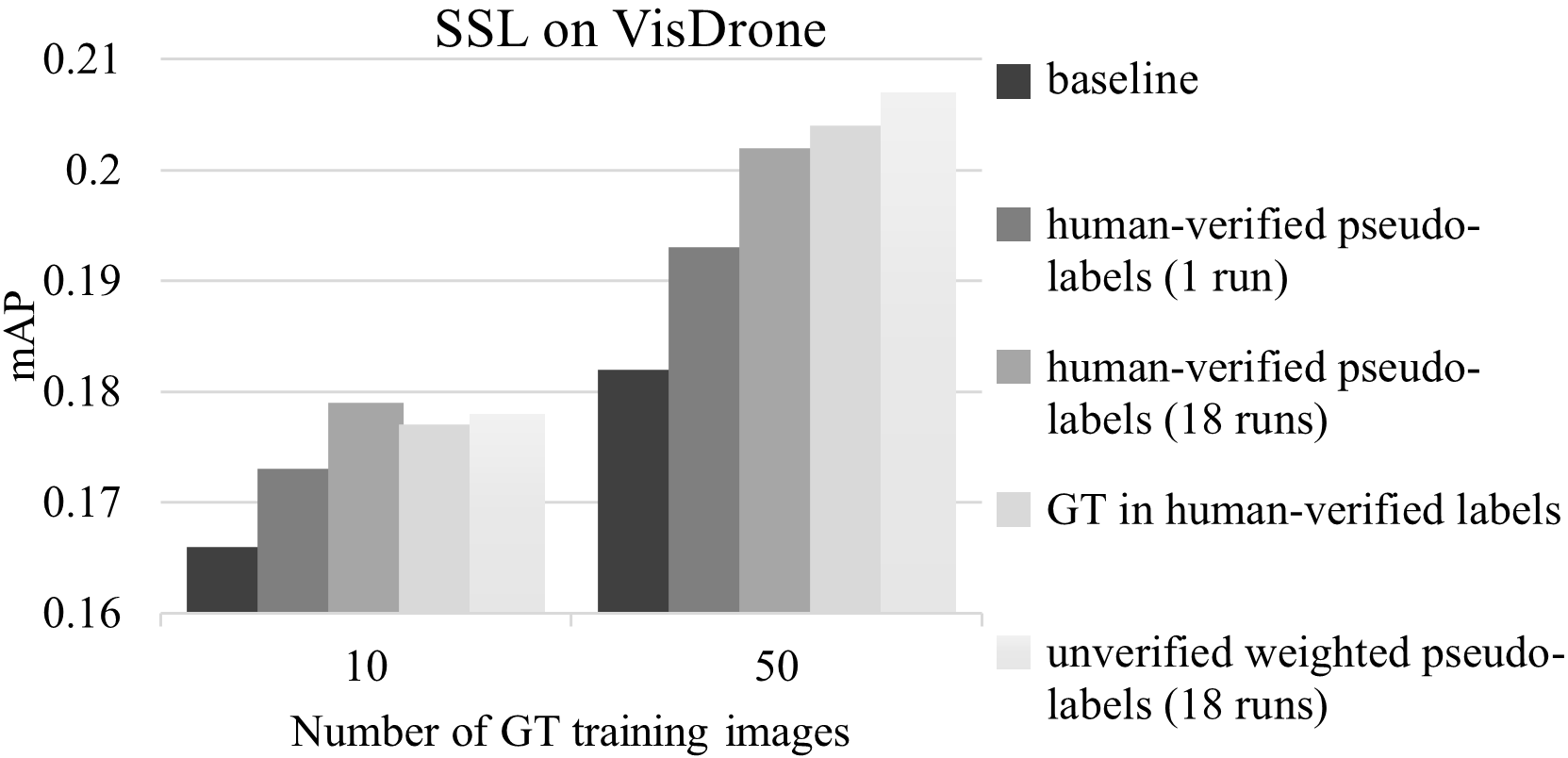}
    \captionof{figure}{Semi-supervised learning is effective for this domain gap, where our weighted pseudo-label loss is a good performer, with an mAP close to or better than human-verified labels, while not requiring any human involvement.}\label{fig:finetuning}
    \end{minipage}
\end{figure}

\begin{figure}[!h]
     \centering
     \begin{subfigure}[b]{0.48\textwidth}
         \centering
         \includegraphics[width=\textwidth]{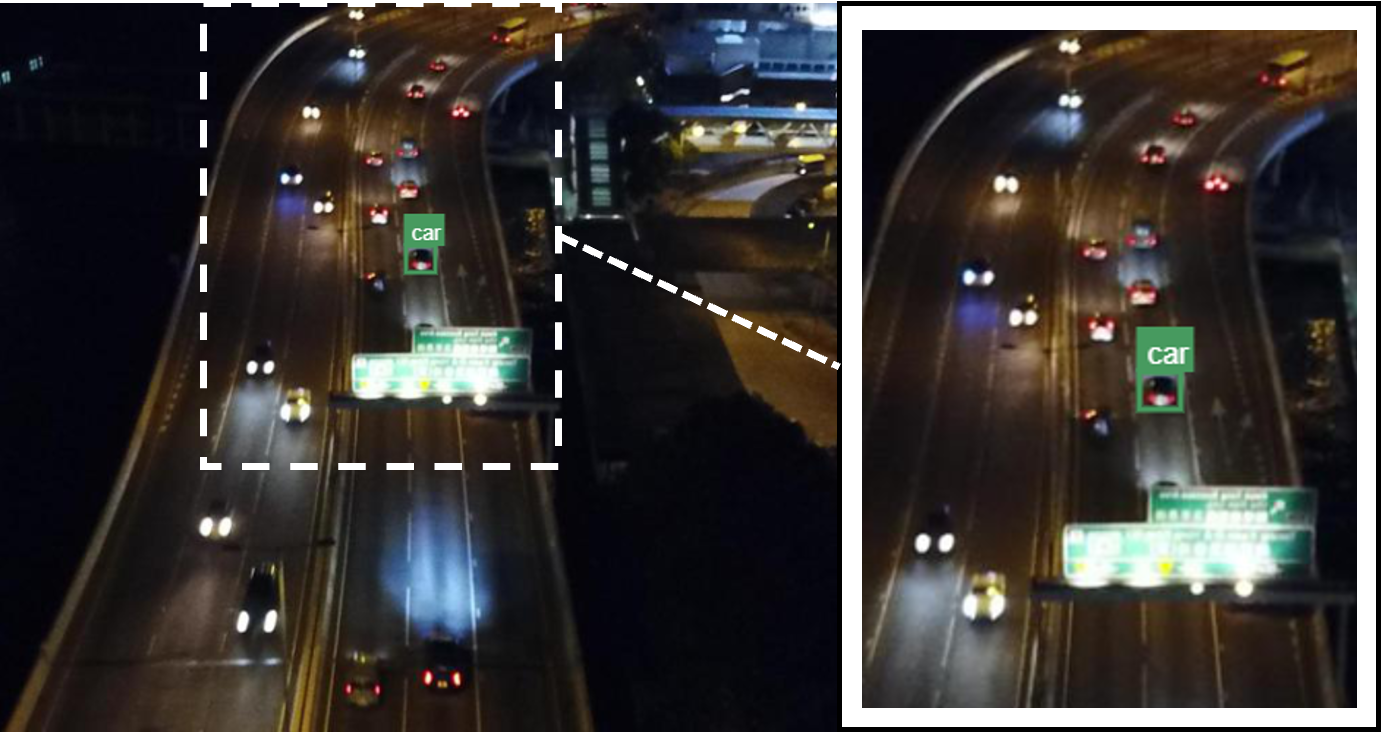}
         \caption{DiffusionDet, standard configuration (1 run, 300 boxes)}
     \end{subfigure}
     \hfill
     \begin{subfigure}[b]{0.48\textwidth}
         \centering
         \includegraphics[width=\textwidth]{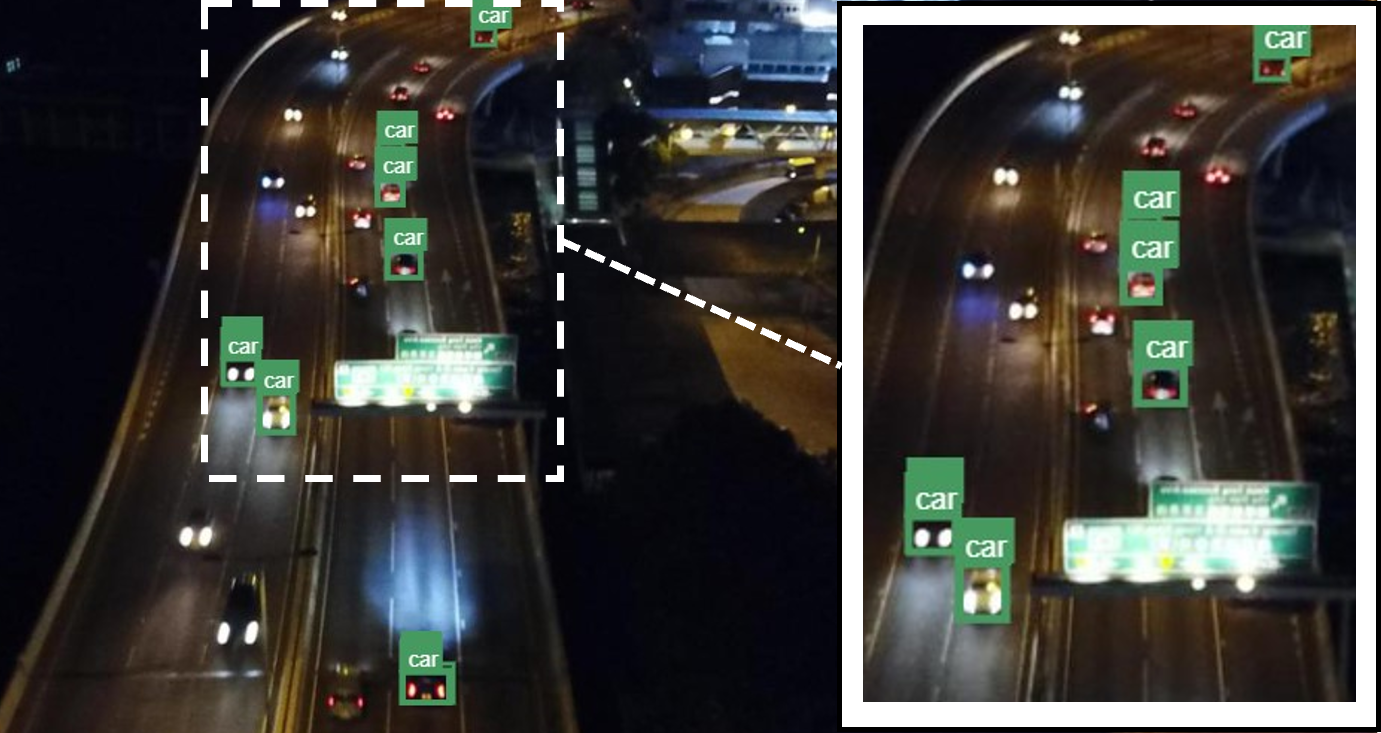}
         \caption{DiffusionDet, 18 runs and finetuned on 50 GT samples}
     \end{subfigure}
  \begin{subfigure}[b]{0.48\textwidth}
         \centering
         \includegraphics[width=\textwidth]{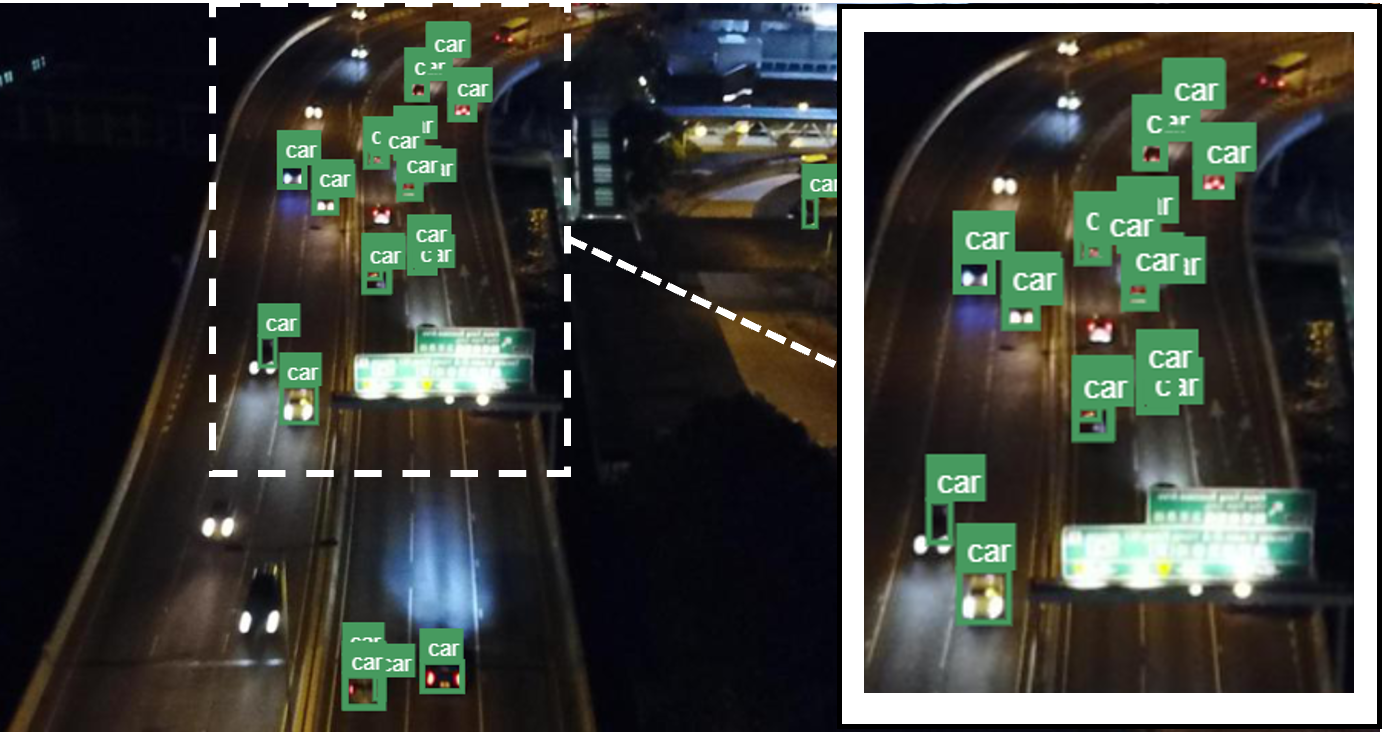}
         \caption{DiffusionDet, 18 runs and finetuned on 50 GT samples in combination with weighted semi-supervised loss}
     \end{subfigure}
     \caption{Improving the finetuning with our weighted semi-supervised loss using pseudo-labels on unlabelled images taking into account their confidences.}
     \label{fig:finetuning_impr}
\end{figure}

Figure \ref{fig:finetuning} shows that the generated pseudo-labels are effective: all models that use them (four bars to the right) outperform the models trained only on the GT images (baseline). This is the case for both 10 and 50 training images. The pseudo-labels after 18 runs with our accumulator are better than a single run of DiffusionDet, as expected, given its improved performance (see Section \ref{sec:exp1}). Surprisingly, the performance of human-verified pseudo-labels is on par with real labels in the same image regions. This is unexpected because the pseudo-labels are imperfect. We have used all pseudo-labels that have a confidence above 0.5. In our weighted loss, the confidence is taken into account during the learning. This strategy is effective: it performs on par (right-most bar) with the human-verified pseudo-labels. Our weighted loss removes the necessity of manual selection, which is labour intensive, without a degradation of performance. For a hard case, we visualise of the improved detections over the standard model, baseline (with finetuning), and the model finetuned with unverified weighted pseudo-labels in Figure \ref{fig:example_data_samples}, the entire picture is shown in Figure \ref{fig:finetuning_impr} (Appendix).

\section{Conclusions}

We leveraged the stochasticity of a diffusion model for object detection, increasing the performance by accumulating variable predictions. For a large domain gap, we have shown that the improved detections on unlabeled images are useful for semi-supervised learning. This is especially the case when weighted according to their confidence with our weighted semi-supervised loss function.

\bibliography{references}

\section*{Appendix}

\end{document}